\documentclass[10pt,twocolumn,letterpaper]{article}

\usepackage{iccv}
\usepackage{times}
\usepackage{epsfig}
\usepackage{graphicx}
\usepackage{amsmath}
\usepackage{amssymb}

\usepackage[breaklinks=true,bookmarks=false]{hyperref}

\iccvfinalcopy % *** Uncomment this line for the final submission

\def\iccvPaperID{****} % *** Enter the ICCV Paper ID here

\ificcvfinal\fi

\begin{document}

%%%%%%%%% TITLE
\title{Strong Instance Segmentation Pipeline for MMSports Challenge}

\author{Bo Yan, Fengliang Qi, Zhuang Li, Yadong Li, Hongbin Wang \\
\small
Ant Group, China \\
}

\maketitle
\ificcvfinal\thispagestyle{empty}\fi

%%%%%%%%% ABSTRACT
\begin{abstract}

The goal of ACM MMSports2022 DeepSportRadar Instance Segmentation Challenge is to tackle the segmentation of individual humans including players, coaches and referees on a basketball court. And the main characteristics of this challenge are there is a high level of occlusions between players and the amount of data is quite limited. In order to address these problems, we designed a strong instance segmentation pipeline. Firstly, we employed a proper data augmentation strategy for this task mainly including photometric distortion transform and copy-paste strategy, which can generate more image instances with a wider distribution. Secondly, we employed a strong segmentation model, Hybrid Task Cascade based detector on the Swin-Base based CBNetV2 backbone, and we add MaskIoU head to HTCMaskHead that can simply and effectively improve the performance of instance segmentation. Finally, the SWA training strategy was applied to improve the performance further. Experimental results demonstrate the proposed pipeline can achieve a competitive result on the DeepSportRadar challenge, with 0.768AP@0.50:0.95 on the challenge set. Source code is available at \url{https://github.com/YJingyu/Instanc_Segmentation_Pro}.

\end{abstract}

%%%%%%%%% BODY TEXT
\section{Introduction}

Instance segmentation applies widely in image editing, image analysis and autonomous driving, etc. Instance segmentation is a fundamental problem in computer vision. Deep learning-based methods have achieved promising results for image instance segmentation over the past few years, such as Mask R-CNN~\cite{he2017mask}, PANet~\cite{liu2018path}, TensorMask~\cite{chen2019tensormask},  CenterMask~\cite{wang2020centermask}, SOLO series~\cite{wang2020solo, wang2020solov2}.

In addition, transformers have made enormous strides in NLP\cite{devlin2018bert, radford2019language}. There are quite a bit of works applying transformers to computer vision\cite{ramachandran2019stand, Zhao_2020_CVPR, carion2020end}, because transformers can capture the non-local and relational nature of images. Especially, Swin Transformer\cite{liu2021swin} has been widely used for many computer vision tasks and achieved successful results, such as detection and segmentation tasks on COCO.

\hfill

In order to address the problems of data-deficient and occlusions, we employed a proper data augmentation strategy mainly including two components, photometric distortion transform and copy-paste strategy, which can increase the diversity of data distribution effectively. And then we employed a strong segmentation model, and add mask scoring~\cite{huang2019mask} to the mask head. Finally, an effective training strategy is applied to improve the model performance further. 

\hfill

%------------------------------------------------------------------------
\section{Approach}

Our approach mainly includes three parts: data augmentation strategy, segmentation model, and training strategy. We first illustrate the data augmentation strategy in Sec.2.1. And the details of segmentation model are presented in Sec.2.2. Finally, the training strategy is introduced in Sec.2.3.

%-------------------------------------------------------------------------
\subsection{Data Augmentation}

For DeepSportRadar instance segmentation, although the scenario seems simple because the task is only to segment players, coaches and referees on images recorded of a basketball court, total images including train, validation and test are also insufficient. In order to generate enough image instances with wider distribution and make the model performs better and more robust, we employed a data augmentation strategy that mainly includes two parts, photometric distortion transform and copy-paste strategy. 

There are four detailed transforms for photometric distortion transform, random brightness, random contrast, random saturation and random hue, every image would be passed through these four transforms. Copy-Paste~\cite{9578639}, which copies objects from one image to another, is particularly useful for instance segmentation. The photometric distortion transform and copy-paste strategy can increase the diversity of data distribution effectively, which can help address the problems of data-deficient and occlusions.

\begin{figure*}[ht]
	\centering
	\includegraphics[scale=0.5]{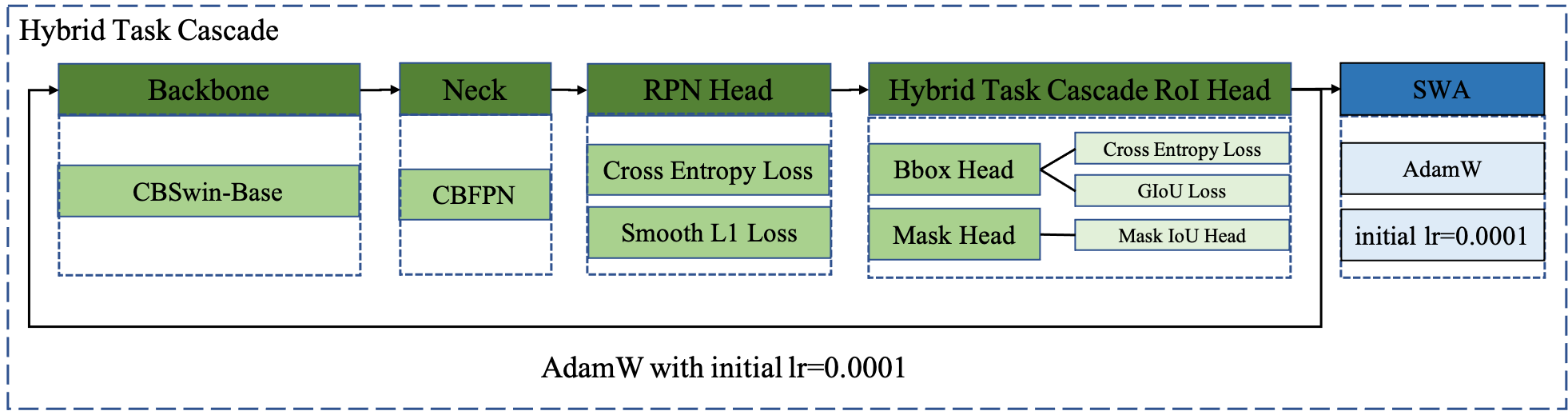}
	\caption{Key components of model architecture and training pipeline for DeepSportRadar challenge.}
	\label{fig1}
\end{figure*}

%-------------------------------------------------------------------------
\subsection{Segmentation Model}

Our segmentation model is Hybrid Task Cascade(HTC)~\cite{chen2019hybrid} based detector on the CBSwin-Base backbone with CBFPN~\cite{Liang2021CBNetV2AC}. The Mask R-CNN with MaskIoU head named as Mask Scoring R-CNN~\cite{huang2019mask}, mask scoring can automatically learn the mask quality instead of relying on the classification confidence of the bounding box, so we add MaskIoU head to HTCMaskHead that can simply and effectively improve the performance of instance segmentation due to the alignment between mask quality and mask score. 

%-------------------------------------------------------------------------
\subsection{Training Strategy}
We first train the model normally with the employed data augmentation strategy. When the model is converged, we use SWA~\cite{zhang2020swa} training strategy to finetune the model, which can make the model better and more robust. The key components of model architecture and training pipeline are shown in Figure \ref{fig1}.

%------------------------------------------------------------------------
\section{Experiments}

%-------------------------------------------------------------------------
\subsection{Training Details}

We use the dataset supplied by DeepSportRadar Instance Segmentation Challenge to train and evaluate the model. In the first stage of training, the pre-trained CBSwin-Base model by COCO is applied, which is supplied by CBNetV2~\cite{Liang2021CBNetV2AC}.

Firstly, we train the model with Adam with decoupled weight decay(AdamW)~\cite{loshchilov2018decoupled} and initial learning rate(lr)=0.0001. When the model is converged, we use SWA training strategy to finetune the model, and the optimizer is also AdamW with initial lr=0.0001. 

The input images are randomly scaled from 820 to 3080 on the short side and up to 3680 on the long side. Then randomly cropped and padded to (1920, 1440). Finally, the data augmentation strategies including random flip, photometric distortion transform, and copy-paste are applied to get the final augmented images, and then input to the segmentation model.

%-------------------------------------------------------------------------
\subsection{Experimental Results}

As shown in Table \ref{tab1}, our proposed pipeline finally achieves 0.768AP@0.50:0.95 on the challenge set of DeepSportRadar Instance Segmentation Challenge.

\begin{table*}[h]
\begin{center}
\resizebox{.95\textwidth}{!}{
\begin{tabular}{llccc}
\hline
Methods & Training Data & Others & mAP-Test & mAP-Challenge\\
\hline
HTC-CBSwinBase & Train+Val & - & 0.734 & - \\
HTC-CBSwinBase+MaskIoU & Train+Val & -  & 0.744 & - \\
HTC-CBSwinBase+MaskIoU & Train+Val & Mask Loss Weight=2.0 & 0.755 & - \\
HTC-CBSwinBase+MaskIoU+SWA & Train+Val & Mask Loss Weight=2.0 & 0.763 & - \\
HTC-CBSwinBase+MaskIoU+SWA+TTA & Train+Val & Mask Loss Weight=2.0 & 0.788 & 0.766 \\
HTC-CBSwinBase+MaskIoU+SWA+TTA & Train+Val+Test & Mask Loss Weight=2.0 & - & 0.768 \\
\hline
\end{tabular}
}
\end{center}
\caption{Experimental results on dataset of DeepSportRadar challenge. \label{tab1}}
\end{table*}
\hfill

\subsection{Ablation Study}

This section elaborates on how we achieve the final result by ablation study to explain our method. 

Our baseline is HTC-CBSwinBase and soft nms~\cite{bodla2017soft} is used for all experiments. The baseline achieves 0.734 mAP on test set. We add MaskIoU head to HTC-CBSwinBase, it can improve the baseline by 0.010 mAP. Then we change the loss weight of mask head from 1.0 to 2.0, it can bring an improvement of 0.011 mAP. And then we use SWA training strategy to finetune the model and it can improve 0.008 mAP. Then test time augmentation(TTA) is applied, our TTA strategy is flip horizontal and multi-scale test with scale factors (1.0, 1.5, 2.0, 2.5, 3.0), TTA is an effective strategy for this task and it can bring an improvement of 0.025 mAP. After the above pipeline, we achieve 0.788 mAP on test set and 0.766 mAP on challenge set.

Finally, we train the model with whole images including training, validation and test dataset, the training settings and epochs are the same as before. We finally achieve 0.768 mAP on the challenge set.

\hfill

%-------------------------------------------------------------------------
\section{Conclusion}

In this paper, we introduce the proposed strong instance segmentation pipeline to address the DeepSportRadar instance segmentation problem. In order to tackle the problems of data-deficient and occlusions, we employed a proper data augmentation strategy that can generate more image instances with a wider distribution. And also we employed a strong segmentation model and made some improvements on it. Finally, an effective training strategy and TTA strategy are introduced. We demonstrate the applicability of proposed pipeline on the DeepSportRadar challenge. Experimental results demonstrate that the proposed pipeline can achieve a competitive result on the ACM MMSports2022 DeepSportRadar Instance Segmentation Challenge, with 0.768AP@0.50:0.95 on the challenge set.

%-------------------------------------------------------------------------

{\small
\bibliographystyle{ieee_fullname}
\bibliography{egbib}
}

\end{document}